\documentclass[a4paper, 10pt, conference]{IEEEtran}
\IEEEoverridecommandlockouts
\usepackage{cite}
\usepackage{amsmath,amssymb,amsfonts}
\usepackage{algorithmic}
\usepackage{graphicx}
\usepackage{textcomp}
\usepackage{xcolor}
\usepackage{multirow}
\usepackage{dblfloatfix}
\def\BibTeX{{\rm B\kern-.05em{\sc i\kern-.025em b}\kern-.08em
    T\kern-.1667em\lower.7ex\hbox{E}\kern-.125emX}}
    
\begin{document}

\title{Hierarchical Visual Localization Based on Sparse Feature Pyramid for Adaptive Reduction of Keypoint Map Size\\
}

\author{\IEEEauthorblockN{
Andrei Potapov,
Mikhail Kurenkov,
Pavel Karpyshev,
Evgeny Yudin, \\
Alena Savinykh,
Evgeny Kruzhkov,
and 
Dzmitry Tsetserukou}
\IEEEauthorblockA{\textit{ISR Laboratory, Skolkovo Institute of Science and Technology, Moscow, Russia}}
\IEEEauthorblockA{$\{$Andrei.Potapov, Mikhail.Kurenkov, Pavel.Karpyshev, Evgeny.Yudin, \\
Alena.Savinykh, Evgeny.Kruzhkov,  D.Tsetserukou$\}$@skoltech.ru}}

\maketitle

\begin{abstract}
Visual localization is a fundamental task for a wide range of applications in the field of robotics. Yet, it is still a complex problem with no universal solution, and the existing approaches are difficult to scale: most state-of-the-art solutions are unable to provide accurate localization without a significant amount of storage space. We propose a hierarchical, low-memory approach to localization based on keypoints with different descriptor lengths. It becomes possible with the use of the developed unsupervised neural network, which predicts a feature pyramid with different descriptor lengths for images. This structure allows applying coarse-to-fine paradigms for localization based on keypoint map, and varying the accuracy of localization by changing the type of the descriptors used in the pipeline. Our approach achieves comparable results in localization accuracy and a significant reduction in memory consumption (up to 16 times) among state-of-the-art methods.
\end{abstract}

\begin{IEEEkeywords}
Visual Localization, Hierarchical Localization, Local Features, Learning-based, Map Size Reduction
\end{IEEEkeywords}

\section{Introduction}
\subsection{Motivation}

The localization of mobile robots is a fundamental and important task, the solution of which is necessary for safe and reliable autonomous operation. The task of localization is aimed at finding the exact position of the robot on a pre-built map. In today’s world, the amount of data stored is constantly growing; the same situation is in mobile robotics --- the size of maps for robot's localization is growing due to the expansion of their implementation areas. The market for autonomous mobile robots maps has shown strong growth over the past few years. It was estimated to be USD 1.4 billion in 2021 and is projected to reach USD 16.9 billion by 2030 \cite{tmirob}. At this stage, the market needs efficient mapping and localization algorithms.


\subsection{Problem Statement}

The key hindrances to achieving high scalability of areas available for autonomous robot operation are the large size of the maps, the high cost of their storage, and the complexity of their updating.
\begin{figure}[!t]
\centerline{\includegraphics[width=8.7 cm]{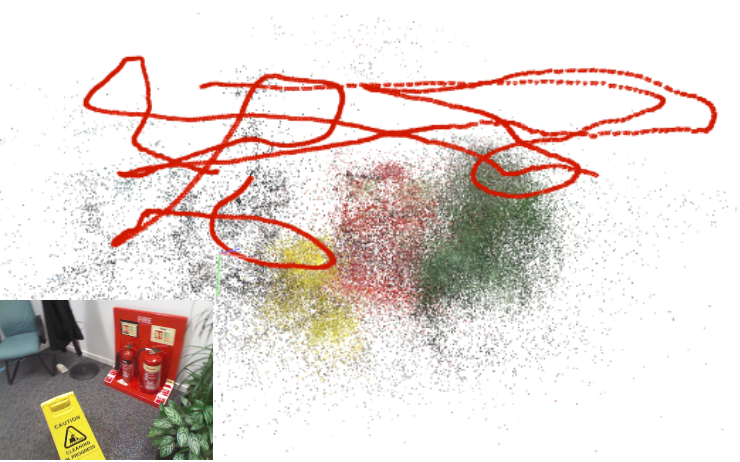}}
\caption{Colmap reconstruction of the Fire scene of the 7-Scenes dataset using the Sparse Feature Pyramid proposed approach}
\label{preview_reconstruction}
\end{figure}

Localization over a large area inevitably leads to an increase in the required memory allocated to the map. A typical model contains around 1 million 3D points for an area of 20,000 m$^2$, which, without compression, would require between 500 MB and 800 MB of storage \cite{germain2021neural}. This makes scaling maps a very complex and costly process. Many modern approaches are aimed at increasing the efficiency of map storage by reducing its size. However, that inevitably results in a loss of useful information and, therefore, degraded accuracy compared to the traditional modern approaches that are not limited in memory. The reduction of the map size is a challenging task not only for outdoor robots \cite{protasov2021cnn, karpyshev2022mucaslam, karpyshev2021autonomous, petrovsky2022two}, but also for indoor application in a large space: for example, robots operating in warehouses \cite{kalinov2019high, kalinov2020warevision, kalinov2021impedance, kalinov2021warevr}, shopping rooms \cite{petrovsky2020customer} and hospitals \cite{perminov2021ultrabot, mikhailovskiy2021ultrabot}. Thus, in this work, we propose an adaptive algorithm that can significantly reduce the amount of memory required for storing a map, and, at the same time, allows adjusting the localization accuracy up to the state-of-the-art results by varying the selected level of the feature pyramid.

\subsection{Related Works}

\textbf{Localization.} Depending on the localization method, the representation of the map used during the operation of the mobile robot differs.
MapNet\cite{brahmbhatt2017mapnet}, inspired by the use of NN \cite{kendall2017geometric, kendall2015posenet} a map for localization is represented by deep neural network (DNN). Its ability to fuse data from different sensors, and time constraints, allows it to increase the accuracy and robustness relative to previous methods \cite{clark2017vidloc,  melekhov2017image, walch2017image}. Despite improvements, these methods still have low localization accuracy and updating the implicit map representation is complex.
On the other hand, T. Sattler et al. \cite{sattler2016efficient} concentrate on improving the localization algorithm, using information from the Structure-from-Motion (SfM) \cite{schonberger2016structure} model for localization. Thanks to the new prioritization scheme and the Active Search algorithm, they manage to achieve highly accurate localization, but the localization model is large, and the localization process takes a long time.

Sarlin et al. \cite{sarlin2019coarse} proposed a hierarchical localization scheme that uses learned features and image retrieval. Global image descriptors reduce the time for keypoint matching. However, a database of images and features still needs to be stored.

The quality of the keypoints used during creation of maps for localization directly affects the accuracy and robustness of the approach. Over time, the number of works aimed at developing the most effective keypoint detectors for solving computer vision problems is constantly increasing. Even decades after the appearance of the first works \cite{forstner1987fast, harris1988combined}, this question remains open.

\textbf{Handcrafted keypoints:} The development of traditional hand-crafted keypoints continues, among which the most famous and successful are Scale-Invariant Feature Transform (SIFT) \cite{ng2003sift}, and ORB (Oriented FAST and Rotated BRIEF) \cite{rublee2011orb}. Based on these approaches, both modifications of methods (SIFT-PCA \cite{juan2009comparison}, DSP-SIFT \cite{dong2015domain}) and completely new, competitive approaches (SURF\cite{bay2006surf}, or DAISY\cite{tola2009daisy}) have been developed, which led to the dominance of methods using local fragment coding images into descriptors and adapting approaches to a variety of computer vision tasks. However, they still have low computation speed and repeatability.


\textbf{Learning-based features:} The emergence of basic learning methods marked a new era in the development of feature extractors. Thus, the work of the FAST detector \cite{rosten2006machine} was aimed at accelerating the process of feature extraction. The works of  MagicPoint\cite{detone2018superpoint}, show a similar ability to match image substructures, but do not detect points of interest. At the same time, LIFT \cite{yi2016lift} (replacement of SIFT by convolutional neural network (CNN)) which contains point of interest detection, orientation estimation and descriptor calculation, but additionally requires control from the classical SfM system and MagicPoint, are among the first successful developments based on the detector. Another approach to feature detection is QuadNetworks \cite{savinov2017quad} which solves the interest point detection problem from an unsupervised approach; however, their system is a patch-based (inputs are small image patches) and a relatively shallow 2-layer network. Hartmann et al.\cite{hartmann2014predicting} propose a slightly different approach for learning to find keypoints: The classifier studies the key points and, based on the points used by the pipeline, it can make key point predictions. However, despite significant acceleration, this approach is still limited by the original keypoint detection algorithm.

\textbf{Map compression.} Also, there are various methods to reduce the size of the map for localization. The most straightforward approach is presented by D. Van et al. \cite{van2018efficient}, where visual information is stored in the form of binary local descriptors that describe the contour of a map point observed by keyframes, which allows this method to be applied to any scenes. However, the reduction in map size is small compared to other state-of-the-art methods.
M. Mera et al. \cite{mera2020efficient} propose to compress the SfM point cloud in order to leave only those points that cover the surface sufficiently for representation, but at the same time provide visual distinctiveness. This approach proposes to compress maps with different score functions, but does not offer a single one suitable for all scenes.
One of the closest works to our method is a paper by F. Camposeco et al. \cite{camposeco2019hybrid}, which proposes manual separation of features, where the first group contains the most useful ones, for which the full length of the descriptor is preserved, and the second group contains compressed points used to further improve localization. This method provides state-of-the-art map compression and localization accuracy metrics, but requires additional processing, which reduces its applicability.

\subsection{Contribution}

\begin{figure}[!t]
\centerline{\includegraphics[width=8.3 cm]{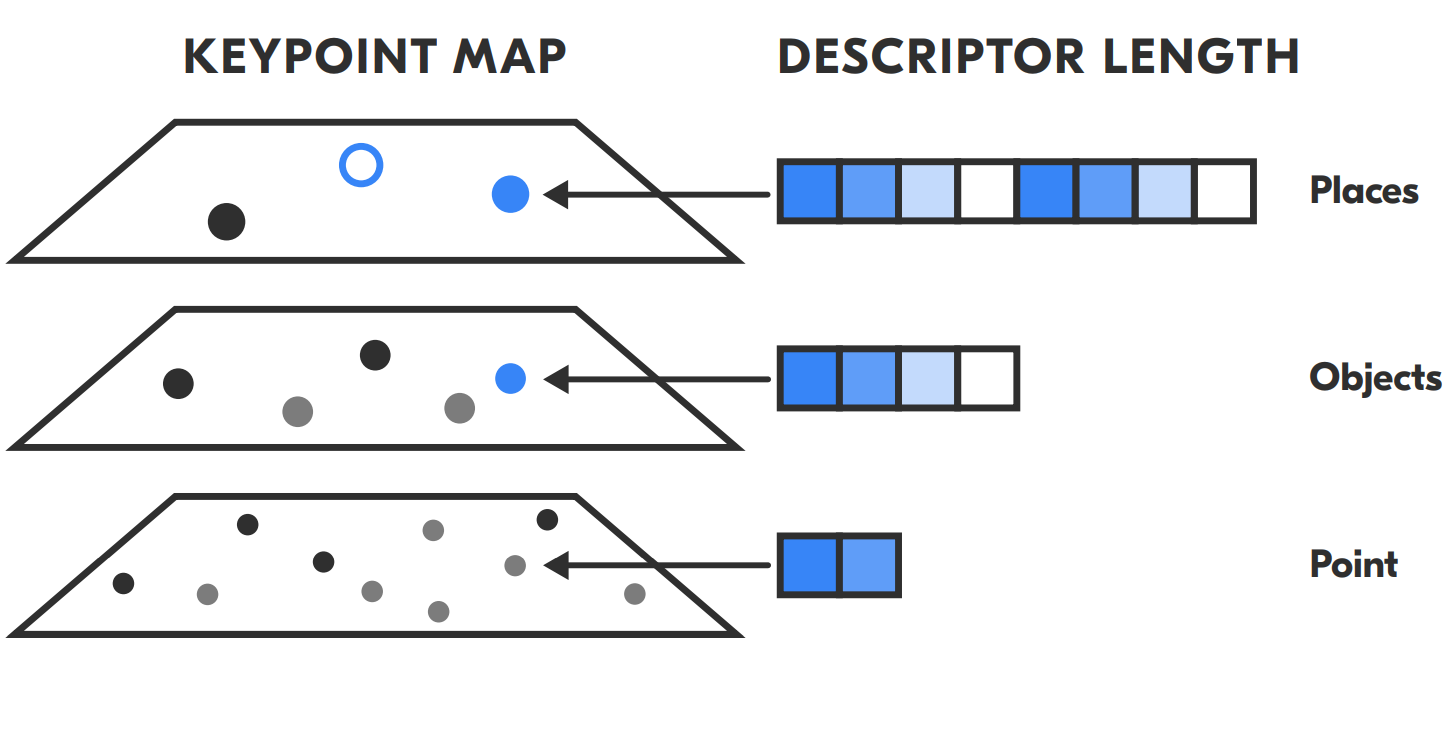}}
\caption{Mapping concept for hierarchical keypoint localization}
\vspace{-1.5em}
\label{mapp}
\end{figure}

\begin{figure} [!b] 
\begin{center}
\includegraphics[width=8.6 cm]{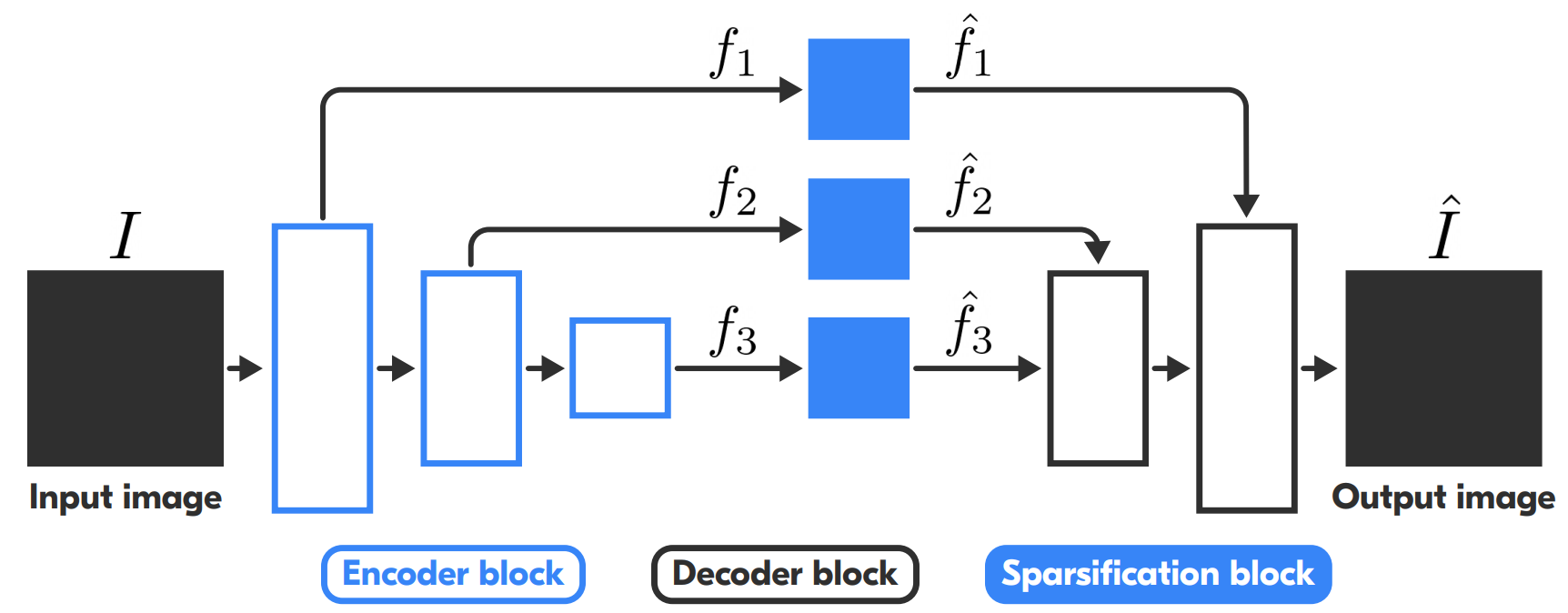}
\caption{The architecture of SPF network: $f$, $\hat{f}$ are the dense feature pyramid, and the sparse feature pyramid, respectively.}
\vspace{-1.5em}
\label{SFP-structure}
\end{center}
\end{figure}

In the scope of this paper, we propose a novel visual localization approach based on sparse feature pyramid, that would allow to significantly decrease the required map (Fig. \ref{preview_reconstruction}) size and provide the possibility of hierarchical localization in coarse-to-fine manner. This is achieved by using a learning-based feature extractor (Fig. \ref{SFP-structure}) able to detect several layers of local features with descriptors of different lengths. These descriptors can be concatenated and combined to achieve higher localization accuracy, or used separately to further decrease the size of the map used in our localization pipeline. The network is trained in an unsupervised manner, which allows applying it to different environments without ground truth data.

\section{Methodology}
In this section, we describe our approach for hierarchical visual localization used for the adaptive localization pipeline that compromises between localization accuracy and map size.

\subsection{Sparse feature pyramid}
To adaptively reduce the size of the memory for storing the descriptor, we introduce sparse feature pyramid neural network, which can predict keypoints with descriptors of different lengths. These keypoints with different descriptor lengths are used for different purposes. Keypoints with long descriptors store information about bigger regions of the scene, and keypoints with short descriptors store information about small details (Fig. \ref{mapp}). 

These keypoints are detected by using describe-to-detect approach. This approach is based on Dense Feature Pyramid \eqref{eq3_1} for keypoint detection. This pyramid consists of several levels of feature images. Deep levels of this pyramid have small heights and width and big dimension size, and shallow levels have big height and width and small feature dimension size. Moreover, for each feature in this feature pyramid, we predict score. We interpret this score function as probability of feature prediction as keypoint. And then by using these probabilities and Monte-Carlo approach we predict sparse feature pyramid as described below.
\begin{equation} \label{eq3_1}
\begin{split}
f_i=F_\theta^{(i)} (I)
\end{split}
\end{equation}
where $I$ is the input image, $F$ is the encoder, $\theta$ is the parameter of neural network (NN), $i$ is the index of level of the feature pyramid.
To make these sparse features descriptive enough, we should have reconstruction ability by using these features. Therefore, we use an autoencoder architecture which encodes an input image to the feature pyramid and then decodes these features to an image which should be equal to the input image. Thus, we should be able to predict the same features on training images under different conditions, including changing the scene's view angle and lightning conditions.

To compromise reconstruction ability and compression ability, we proposed use of loss function \eqref{eq3_6} which is the sum of reconstruction loss and compression loss. Reconstruction loss is a L2 norm between the input and output images. And compression loss is the sum of predicted scores for each image multiplied by feature dimension. This compression loss is an average size of compression of the input image to feature pyramid.
\begin{equation} \label{eq3_6}
\begin{split}
L=(I-\hat{I})_1 + \lambda\displaystyle\sum_{i=1} ^{n} \sum_{j}p_{i,j}*d_{i,j}
\end{split}
\end{equation}
where $I$ is the input image, $\hat{I}$ is the output image, $p_{i,j}$ is the probabilistic mask, $\lambda$ and $d_{i,j}$ is the loss function weights.

However, to use reconstruction loss with sparse keypoints, we need to use an attention mask, which can highlight keypoints based on the feature pyramid. If this mask is hard-coded with threshold, this mask is not differentiable, and, as a result, this mask can not be trained by gradient decent. Therefore, we use a stochastic mask with scores as probabilities to be either one or zero \eqref{eq3_2}, \eqref{eq3_3}, and with Bernoulli reparametrization trick for differentiation. 
\begin{equation} \label{eq3_2}
\begin{split}
p_{i,j}=\sigma(\omega_{i,j}*f_{i,j})
\end{split}
\end{equation}
\begin{equation} \label{eq3_3}
\begin{split}
m_{i,j}{\sim}Bernoulli(p_{i,j})
\end{split}
\end{equation}
where $\sigma$ is the sigmoid function, $\omega_{i,j}$ is the parameter of the neural network.

Multiplying the binary mask by the Dense Feature Pyramid, we get the Sparse Feature Pyramid \eqref{eq3_4}. Features with non-zero values are interpreted as input image keypoints, whose centers are in the pixel, predicted at the point and whose descriptors are features in the given pixel. Thus, from different layers of this sparse feature pyramid, we receive keypoints with different descriptor length. 
\begin{figure} [!ht] 
\begin{center}
\includegraphics[width=8.3 cm]{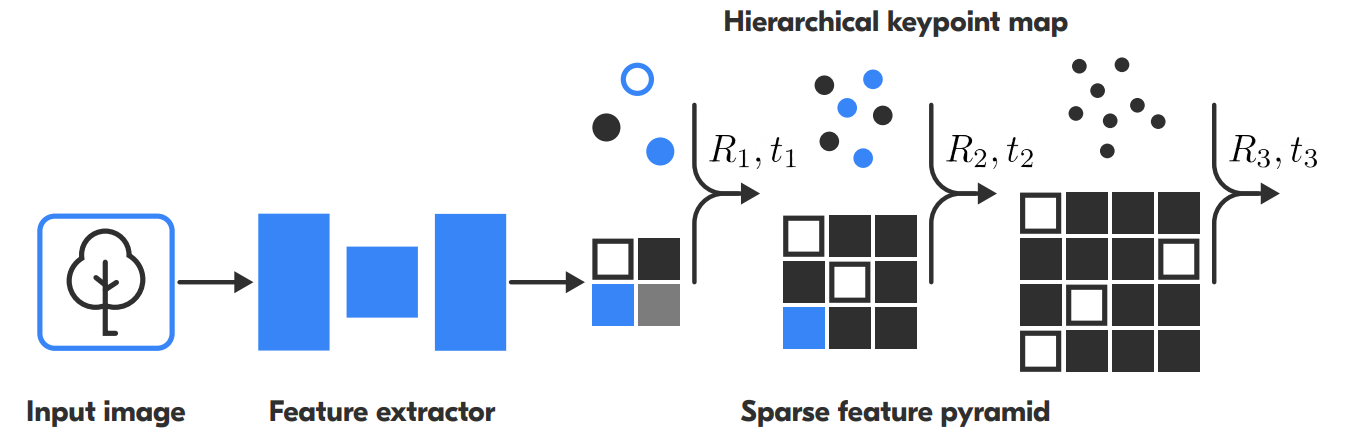}
\caption{Hierarchical localization pipeline from coarse to fine. $R,t\in SE(3)$ is camera position, $R_1,t_1$ is coarse position, $R_3,t_3$ is fine position.}
\vspace{-1.5em}
\label{loc_pp}
\end{center}
\end{figure}
\begin{table*}[!hb]
\caption{Median translation error (m) and rotation error ($^{\circ}$) for various methods on the 7-Scenes dataset}
\resizebox{\textwidth}{!}{%
\begin{tabular}{llcccccccccccccc}
\hline
\multicolumn{2}{l}{\multirow{2}{*}{Method}} &
  \multicolumn{2}{c}{Chess} &
  \multicolumn{2}{c}{Fire} &
  \multicolumn{2}{c}{Heads} &
  \multicolumn{2}{c}{Office} &
  \multicolumn{2}{c}{Pumpkin} &
  \multicolumn{2}{c}{Kitchen} &
  \multicolumn{2}{c}{Stairs} \\
\multicolumn{2}{l}{} &
  \shortstack{MB \\ used} &
  \shortstack{Errors\\ {[}cm, $^{\circ}${]}} &
  \shortstack{MB \\ used} &
  \shortstack {Errors\\ {[}cm, $^{\circ}${]}} &
  \shortstack{MB \\ used} &
  \shortstack {Errors\\ {[}cm, $^{\circ}${]}} &
  \shortstack{MB \\ used} &
  \shortstack {Errors\\ {[}cm, $^{\circ}${]}} &
  \shortstack{MB \\ used} &
  \shortstack {Errors\\ {[}cm, $^{\circ}${]}} &
  \shortstack{MB \\ used} &
  \shortstack {Errors\\ {[}cm, $^{\circ}${]}} &
  \shortstack{MB \\ used} &
  \shortstack {Errors\\ {[}cm, $^{\circ}${]}} \\ \hline
\multicolumn{2}{l}{Active Search\cite{sattler2016efficient}} & $>$100 & 3/0.87 & $>$100 & 2/1.01 & $>$100 & 1/0.82	 & $>$100 & 4/1.15 & $>$100 & 7/1.69 & $>$100 & 5/1.72 & $>$100 & 4/1.01 \\
\multicolumn{2}{l}{DSAC*\cite{brachmann2018learning}}                & 28 & 2/1.10 & 28 & 2/1.24 & 28 & 1/1.82 & 28 & 3/1.15 & 28 & 4/1.34 & 28 & 4/1.68	 & 28 & 3/1.16 \\
\multicolumn{2}{l}{SuperPoint+SuperGlue\cite{sarlin2019coarse}} & $>>$100 & 2/0.84 & $>>$100 & 2/0.93 & $>>$100 & 1/0.74 & $>>$100 & 3/0.92 & $>>$100 & 5/1.27 & $>>$100 & 4/1.40 & $>>$100 & 5/1.47 \\
\multicolumn{2}{l}{PoseNet\cite{kendall2015posenet}}                 & 50 &  13/4.48 & 50 & 27/11.3 & 50 & 17/13.0 & 50 & 19/5.55 & 50 & 26/4.75 & 50 & 23/5.35 & 50 & 35/12.4 \\ \hline

\multicolumn{2}{l}  {SFP map lvl2 fulld, loc lvl2 fulld} & 48.64 & 3.6/1.13 & 21.50 & 4.2/1.64 & 9.6 & 2.3/1.58 & 45.7 & 6.2/1.65 & 25.6 & 8.3/2.26 & 60.16 & 6.2/1.8 & 11.9 & 7.1/1.93 \\
\multicolumn{2}{l}{SFP map lvl2 fulld, loc lvl2 shortd}          & 24.32 & 3.3/1.02 & 10.75 & 3.8/1.5 & 4.8 & 2.2/1.43 & 22.84 & 5.8/1.49 & 12.8 & 8.1/2.15 & 30.08 & 6.1/1.69 & 5.95 & 6.9/1.85 \\
\multicolumn{2}{l}{SFP map lvl2 shortd, loc lvl2 shortd}          & 24.32 & 3.1/0.95 & 10.75 & 3.5/1.37 & 4.8 & 1.9/1.38 & 22.84 & 5.6/1.36 & 12.8 & 8/2.07 & 30.08 & 5.9/1.57 & 5.95 & 6.8/1.82 \\ \hline
\end{tabular}%
}
\label{7scenes-loc}
\end{table*}

\begin{equation} \label{eq3_4}
\begin{split}
\hat{f}_{i,j}=m_{i,j}*f_{i,j}
\end{split}
\end{equation}
Summing up, the overall structure of the pipeline is as follows: we use a CNN autoencoder based on the U-Net architecture, that aims to predict sets of sparse features in images. The standard U-NET architecture uses two types of blocks: encoder and decoder. In our implementation, the encoder block includes successive executions of the MaxPool 2D operation (Kernel size=3, stride=2 and padding=1), batch normalization 2D, ReLU, Convolution 2D. The decoder block consists of successive Upsample Convolution and Convolution operations with ReLU activation function. The main modification of U-Net based autoencoder is the introduction of a sparsification block into the skip connections between the layers of the CNN. Thus, we make sparse feature pyramid for prediction keypoint with different descriptor length

\begin{figure*} [!ht] 
\begin{center}
\includegraphics[width=13.8 cm]{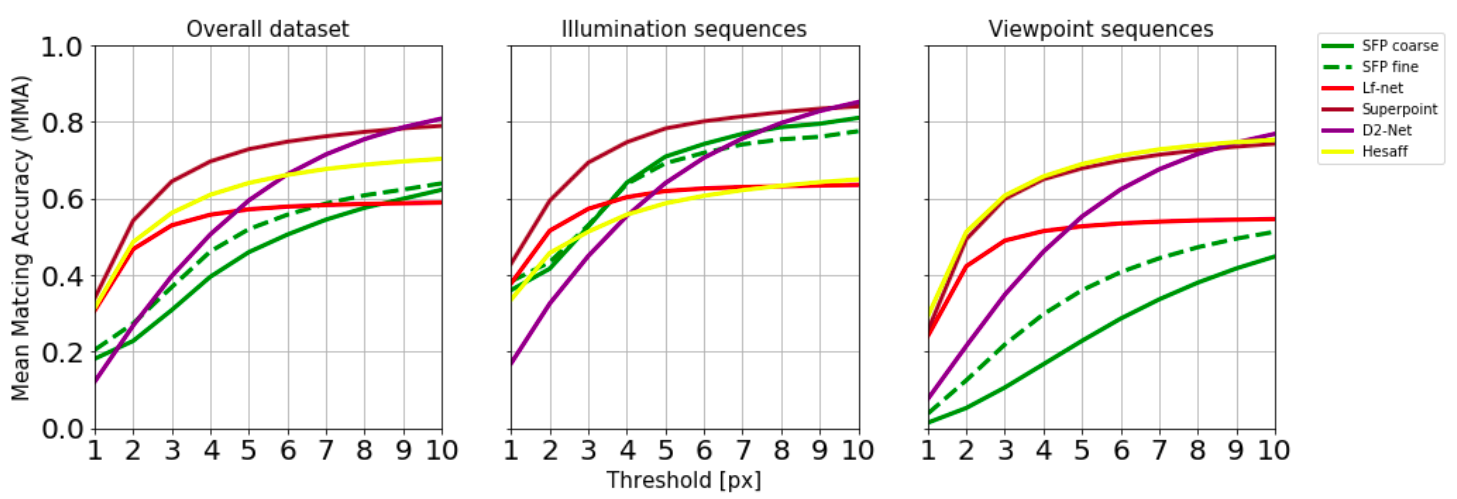}
\caption{Comparison of Mean Matching Accuracy on the HPatches dataset. The x-axis indicates threshold in pixels, the y-axis indicates Mean Matching Accuracy.}
\vspace{-1.5em}
\label{mma}
\end{center}
\end{figure*}

\subsection{Hierarchical keypoint localization}

We propose to use hierarchical approach for localization (Fig. \ref{loc_pp}) with sparse feature pyramid. This approach is based on coarse-to-fine manner for localization.

The first step of localization is getting a pair of images by matching the global descriptors received from NetVLAD. After that, we use different layers of keypoints obtained by Sparse Feature Pyramid (SFP). Starting from a deeper layer, we match them to map keypoints to detect a more accurate position. Then, by using this position as initial position, we use reprojection distances for next keypoint matching. Keypoint matching is based on the scalar product of descriptors. Also, we filter keypoint matches with high reprojection error to speed up the matching process and reduce false positive matches.

For map building, we concatenate the descriptors from all layers of the NN. However, for localization, we use keypoints from only specific layers. For sub-pixel approximation of descriptors, we use bilinear interpolation. For more accurate keypoint prediction, we use interpolation with non-maximal suppression, removing closely spaced keypoints.




\section{Experiments and Results}
\begin{table*}[!ht]

\caption{Median translation error (m) and rotation error ($^{\circ}$) on the Cambridge Landmarks dataset.}
\vspace{2pt}
\resizebox{\textwidth}{!}{%
\begin{tabular}{llcccccccccccccc}
\hline
\multicolumn{2}{l}{\multirow{2}{*}{Method}} &
  \multicolumn{2}{c}{Kings College} &
  \multicolumn{2}{c}{Old Hospital} &
  \multicolumn{2}{c}{Shop Facade} &
  \multicolumn{2}{c}{St. Mary's Church} \\
\multicolumn{2}{l}{} &
  \shortstack{MB \\ used} &
  \shortstack{Errors\\ {[}m, $^{\circ}${]}} &
  \shortstack{MB \\ used} &
  \shortstack {Errors\\ {[}m, $^{\circ}${]}} &
  \shortstack{MB \\ used} &
  \shortstack {Errors\\ {[}m, $^{\circ}${]}} &
  \shortstack{MB \\ used} &
  \shortstack {Errors\\ {[}m, $^{\circ}${]}} \\ \hline
\multicolumn{2}{l}{KCP} & 5.9 & 0.99/0.86 & 8.2 & 1.19/1.00 & 1.3 & 0.44/0.80 & 24 & 0.40/0.61 \\
\multicolumn{2}{l}{Hybrid comp.} & 1.01 & 0.81/0.59 & 0.62 & 0.75/1.01 & 0.16 & 0.19/0.54 & 1.34 & 0.50/0.49 \\
\multicolumn{2}{l}{Efficient comp.} & 2.2 & 0.92/0.59 & 1.1 & 1.24/0.96 & 0.41 & 0.31/0.59 & 3.3 & 0.43/0.64  \\
\multicolumn{2}{l}{DenseVlad} & 10.06 & 2.80/5.72 & 13.98 & 4.01/7.13 & 3.61 & 1.11/7.61 & 23.23 & 2.21/8.00\\
\multicolumn{2}{l}{PoseNet} & 50 & 1.92/5.40 & 50 & 2.31/5.38 & 50 & 1.46/8.08 & 50 & 2.65/8.48\\
\multicolumn{2}{l}{DSAC++} & 207 &  0.18/0.30 & 207 & 0.20/0.30 & 207 & 0.06/0.30 & 207 & 0.13/0.40\\
\multicolumn{2}{l}{Active Search} & 275 &  0.57/0.70 & 140 & 0.52/1.12 & 38.7 & 0.12/0.41 & 359 & 0.22/0.62 \\\hline

\multicolumn{2}{l}{SFP map lvl2 fulld, loc lvl2 fulld} & 3.3 & 0.47/0.68 & 2.6 & 0.39/0.78 & 0.4 & 0.24/0.62 & 3.4 & 0.44/1.17 \\
\multicolumn{2}{l}{SFP map lvl2 fulld, loc lvl2 shortd} & 1.65 & 0.36/0.59 & 1.3 & 0.35/0.71 & 0.2 & 0.16/0.57 & 1.7 & 0.38/1.09 \\
\multicolumn{2}{l}{SFP map lvl2 shortd, loc lvl2 shortd} & 1.65 & 0.33/0.54 & 1.3 & 0.3/0.68 & 0.2 & 0.11/0.53 & 1.7 & 0.29/1.00 \\ \hline

\end{tabular}%
}
\label{Cambridge-loc-main}
\end{table*}

In order to evaluate the proposed approach, two sets of experiments were conducted. The first set was aimed at assessing the quality of keypoints obtained using the proposed neural network and comparing them with features obtained using state-of-the-art methods. The second set was aimed at assessing the localization accuracy and map size of the proposed method.

\subsection{Local features evaluation}
We started the evaluation of our method by verifying the quality of the features extracted by the neural network on the HPatches \cite{balntas2017hpatches} dataset.

\textbf{Datasets.}
We evaluated the quality of our keypoints obtained with SFP using the HPatches dataset. It consists of 116 sequences with changes in lighting (with 57 sequences) or viewpoints (with 59 sequences) conditions, and it contains image transformation matrices as a ground truth for establishing correspondences between 2D images of the dataset. The first image in each sequence is used as a reference, and the subsequent ones form pairs with it, with increasing complexity of viewing angle and lighting change.

\textbf{Metrics.}
Keypoint detection and description was performed for each dataset image, after which matching was performed using the nearest neighbor search. The correctness of the matches was checked using the transformation matrix, a match is considered correct if the reprojection error (distance between the keypoints after reprojection) does not exceed the threshold in pixels. Next, as was done in \cite{revaud2019r2d2, dusmanu2019d2}, Mean Matching Accuracy (MMA) was calculated, displaying the mean percentage of correct matches for each pair of images in sequence overall dataset for each of the thresholds.

\textbf{Implementation details.}
When assessing the quality of keypoints, we used various combinations of parameters. We varied the layers from which we took keypoints, combined them with each other, and also used descriptors of different lengths. In addition to this, we have tested various scenarios such as changing the required score to use a keypoint, keypoint mask interpolation, and non-maximal suppression.

\textbf{Results.} In the case of using the full length descriptors, there is a decrease in accuracy on sequences with significant changes in the viewing angle, since the coarse part of the descriptor can only approximately estimate the viewing angle. The use of bilinear interpolation and non-maximal suppression allows us to obtain a smooth and sparse picture, which improves the accuracy of the match.

Fig. \ref{mma} shows a comparison of the most successful combinations of initial conditions. Thus, the use of long descriptors in combination with interpolation and non-maximal suppression (green line) allows to achieve the best results on illumination sequences, surpassing LF-Net\cite{ono2018lf} and HesAff\cite{mishchuk2017working} by more than 15 $\%$, and at the same time only slightly (less than 5$\%$) yielding to SuperPoint\cite{detone2018superpoint} and D2-Net\cite{Dusmanu2019CVPR}. Using only short (green dashed line) descriptors  with the same conditions, the most accurate results can be obtained for sequences with viewpoint changes.
\par

\subsection{Localization}
Based on the results obtained in the previous experiment, we present the results of map localization and compression on the 7-Scenes \cite{shotton2013scene} and Cambridge Landmarks datasets.

\textbf{Datasets.}
Both datasets contain scenes, each of which is divided into training and validation sets, containing an ordered set of color images, depth information, and camera positions in homogeneous coordinates. A set of training images was used to compile the SfM model, and a set of validation images was used for localization.

\textbf{Metrics.}
For a quantitative comparison of localization, the median errors for camera translation and rotation were calculated, as well as the memory costs for storing the map for localization.

\textbf{Implementation details.}
We followed Sarlin et al.\cite{sarlin2019coarse} proposed localization pipeline in our experiment. To build the SfM model, features from each hierarchical level with the full length of the descriptor were used. Localization took place in several stages. At the first stage of localization, we received a couple of images using NetVLAD\cite{arandjelovic2016netvlad}, which based on the Vector of Locally Aggregated Descriptors (VLAD) method for instance search and image classification. This CNN framework works in an end-to-end manner, and is directly optimized for the task of place recognition. It allows to extract a fixed-size vector representation of an image in order to perform the visual search by finding the nearest database image to the query, during localization. After which we matched local features of various levels of sparse feature pyramids step by step. Moving from a deeper layer, we gradually increased the accuracy of localization, up to using small features that describe points in the environment.

\setlength{\parskip}{0pt}
\textbf{Results.} For the indoor scenario the most precise version is able to achieve a localization average accuracy of 4.97/1.5 [cm, $\deg$], as opposed to DSAC*\cite{brachmann2018learning}, Active Search and combination of SuperPoint+SuperGlue with 2.71/1.35, 3.71/1.18, 3.14/1.08 [cm, $\deg$] respectively. At the same time, the size of the map generated by the proposed pipeline was 16 times smaller than the common map of learning-based methods and, the average memory consumption was 15.93 MB, which is less than DSAC* and PoseNet with map size 28 and 50 MB respectively.

For outdoor scenario the most accurate version is able to achieve an average accuracy of 0.26/0.69 [cm, $\deg$], which is the most accurate method except for DSAC++, which, with an accuracy of 0.14/0.32 [cm, $\deg$], has a map weight 184 times more (1.12 and 207 MB, respectively). At the same time, the only compression method presented, that has commensurate metrics of map weight and localization accuracy is Hybrid Compression. The average map weight of the proposed approach is slightly larger (1.12 vs 0.78 MB), but the translation and rotation error localization accuracy is twice as high (0.26/0.69 vs 0.56/0.66 [cm, $\deg$]). Also, Hybrid Compression is a supervised method, which makes it impossible to work without ground truth.

\section{Conclusion and Discussion}
In the scope of this research, we proposed a novel hierarchical approach to visual localization, using image descriptors with different lengths to reduce map size without sacrificing localization accuracy. Our sparse feature pyramid-based feature extraction pipeline extracts multiple levels of features with varying descriptor length. Our approach was evaluated in two experiments.
\par
The first set has shown that the proposed descriptors are descriptive enough to reconstruct the original images, and, thus, can be used in visual localization.  In the second set of experiments, the proposed pipeline was tested on outdoor and indoor datasets in terms of localization accuracy and resulting map size, and compared to modern state-of-the-art approaches.
\par
According to the results on the outdoor Cambridge Landmarks dataset, proposed approach outperforms all compression methods in the localization accuracy metric (0.26/0.69 [cm, $\deg$] for proposed approach and 0.56/0.66 [cm, $\deg$] for the second-best compression method) and slightly yielded in average map size to the most successful Hybrid Compression method (1.12 and 0.78 MB, respectively).
\par
In our future work, we plan to adapt and expand the proposed algorithm to increase its applicability in real-world applications

\bibliographystyle{IEEEtran}
\bibliography{literature}

\begin{thebibliography}{10}
\providecommand{\url}[1]{#1}
\csname url@samestyle\endcsname
\providecommand{\newblock}{\relax}
\providecommand{\bibinfo}[2]{#2}
\providecommand{\BIBentrySTDinterwordspacing}{\spaceskip=0pt\relax}
\providecommand{\BIBentryALTinterwordstretchfactor}{4}
\providecommand{\BIBentryALTinterwordspacing}{\spaceskip=\fontdimen2\font plus
\BIBentryALTinterwordstretchfactor\fontdimen3\font minus
  \fontdimen4\font\relax}
\providecommand{\BIBforeignlanguage}[2]{{%
\expandafter\ifx\csname l@#1\endcsname\relax
\typeout{** WARNING: IEEEtran.bst: No hyphenation pattern has been}%
\typeout{** loaded for the language `#1'. Using the pattern for}%
\typeout{** the default language instead.}%
\else
\language=\csname l@#1\endcsname
\fi
#2}}
\providecommand{\BIBdecl}{\relax}
\BIBdecl

\bibitem{tmirob}
\BIBentryALTinterwordspacing
``{Autonomous Mobile Robot Market by Type, by Application, by End-User - Global
  Opportunity Analysis and Industry Forecast 2022-2030},'' 2022, [accessed
  28-April-2022]. [Online]. Available:
  \url{https://www.researchandmarkets.com/reports/5529480/autonomous-mobile-robot-market-by-type-by/}
\BIBentrySTDinterwordspacing

\bibitem{germain2021neural}
H.~Germain, V.~Lepetit, and G.~Bourmaud, ``Neural reprojection error: Merging
  feature learning and camera pose estimation,'' in \emph{Proceedings of the
  IEEE/CVF Conference on Computer Vision and Pattern Recognition}, 2021, pp.
  414--423.

\bibitem{protasov2021cnn}
S.~Protasov, P.~Karpyshev, I.~Kalinov, P.~Kopanev, N.~Mikhailovskiy,
  A.~Sedunin, and D.~Tsetserukou, ``Cnn-based omnidirectional object detection
  for hermesbot autonomous delivery robot with preliminary frame
  classification,'' in \emph{2021 20th International Conference on Advanced
  Robotics (ICAR)}.\hskip 1em plus 0.5em minus 0.4em\relax IEEE, 2021, pp.
  517--522.

\bibitem{karpyshev2022mucaslam}
P.~Karpyshev, E.~Kruzhkov, E.~Yudin, A.~Savinykh, A.~Potapov, M.~Kurenkov,
  A.~Kolomeytsev, I.~Kalinov, and D.~Tsetserukou, ``Mucaslam: Cnn-based frame
  quality assessment for mobile robot with omnidirectional visual slam,'' in
  \emph{2022 IEEE 18th International Conference on Automation Science and
  Engineering (CASE)}.\hskip 1em plus 0.5em minus 0.4em\relax IEEE, 2022, pp.
  368--373.

\bibitem{karpyshev2021autonomous}
P.~Karpyshev, V.~Ilin, I.~Kalinov, A.~Petrovsky, and D.~Tsetserukou,
  ``Autonomous mobile robot for apple plant disease detection based on cnn and
  multi-spectral vision system,'' in \emph{2021 IEEE/SICE International
  Symposium on System Integration (SII)}.\hskip 1em plus 0.5em minus
  0.4em\relax IEEE, 2021, pp. 157--162.

\bibitem{petrovsky2022two}
A.~Petrovsky, I.~Kalinov, P.~Karpyshev, D.~Tsetserukou, A.~Ivanov, and
  A.~Golkar, ``The two-wheeled robotic swarm concept for mars exploration,''
  \emph{Acta Astronautica}, vol. 194, pp. 1--8, 2022.

\bibitem{kalinov2019high}
I.~Kalinov, E.~Safronov, R.~Agishev, M.~Kurenkov, and D.~Tsetserukou,
  ``High-precision uav localization system for landing on a mobile
  collaborative robot based on an ir marker pattern recognition,'' in
  \emph{2019 IEEE 89th Vehicular Technology Conference (VTC2019-Spring)}.\hskip
  1em plus 0.5em minus 0.4em\relax IEEE, 2019, pp. 1--6.

\bibitem{kalinov2020warevision}
I.~Kalinov, A.~Petrovsky, V.~Ilin, E.~Pristanskiy, M.~Kurenkov, V.~Ramzhaev,
  I.~Idrisov, and D.~Tsetserukou, ``Warevision: Cnn barcode detection-based uav
  trajectory optimization for autonomous warehouse stocktaking,'' \emph{IEEE
  Robotics and Automation Letters}, vol.~5, no.~4, pp. 6647--6653, 2020.

\bibitem{kalinov2021impedance}
I.~Kalinov, A.~Petrovsky, R.~Agishev, P.~Karpyshev, and D.~Tsetserukou,
  ``Impedance-based control for soft uav landing on a ground robot in
  heterogeneous robotic system,'' in \emph{2021 International Conference on
  Unmanned Aircraft Systems (ICUAS)}.\hskip 1em plus 0.5em minus 0.4em\relax
  IEEE, 2021, pp. 1653--1658.

\bibitem{kalinov2021warevr}
I.~Kalinov, D.~Trinitatova, and D.~Tsetserukou, ``Warevr: Virtual reality
  interface for supervision of autonomous robotic system aimed at warehouse
  stocktaking,'' in \emph{2021 ieee international conference on systems, man,
  and cybernetics (smc)}.\hskip 1em plus 0.5em minus 0.4em\relax IEEE, 2021,
  pp. 2139--2145.

\bibitem{petrovsky2020customer}
A.~Petrovsky, I.~Kalinov, P.~Karpyshev, M.~Kurenkov, V.~Ramzhaev, V.~Ilin, and
  D.~Tsetserukou, ``Customer behavior analytics using an autonomous
  robotics-based system,'' in \emph{2020 16th International Conference on
  Control, Automation, Robotics and Vision (ICARCV)}.\hskip 1em plus 0.5em
  minus 0.4em\relax IEEE, 2020, pp. 327--332.

\bibitem{perminov2021ultrabot}
S.~Perminov, N.~Mikhailovskiy, A.~Sedunin, I.~Okunevich, I.~Kalinov,
  M.~Kurenkov, and D.~Tsetserukou, ``Ultrabot: Autonomous mobile robot for
  indoor uv-c disinfection,'' in \emph{2021 IEEE 17th International Conference
  on Automation Science and Engineering (CASE)}.\hskip 1em plus 0.5em minus
  0.4em\relax IEEE, 2021, pp. 2147--2152.

\bibitem{mikhailovskiy2021ultrabot}
N.~Mikhailovskiy, A.~Sedunin, S.~Perminov, I.~Kalinov, and D.~Tsetserukou,
  ``Ultrabot: Autonomous mobile robot for indoor uv-c disinfection with
  non-trivial shape of disinfection zone,'' in \emph{2021 26th IEEE
  International Conference on Emerging Technologies and Factory Automation
  (ETFA)}.\hskip 1em plus 0.5em minus 0.4em\relax IEEE, 2021, pp. 1--7.

\bibitem{brahmbhatt2017mapnet}
S.~Brahmbhatt, J.~Gu, K.~Kim, J.~Hays, and J.~Kautz, ``Mapnet: Geometry-aware
  learning of maps for camera localization,'' 2017.

\bibitem{kendall2017geometric}
A.~Kendall and R.~Cipolla, ``Geometric loss functions for camera pose
  regression with deep learning,'' in \emph{Proceedings of the IEEE conference
  on computer vision and pattern recognition}, 2017, pp. 5974--5983.

\bibitem{kendall2015posenet}
A.~Kendall, M.~Grimes, and R.~Cipolla, ``Posenet: A convolutional network for
  real-time 6-dof camera relocalization,'' in \emph{Proceedings of the IEEE
  international conference on computer vision}, 2015, pp. 2938--2946.

\bibitem{clark2017vidloc}
R.~Clark, S.~Wang, A.~Markham, N.~Trigoni, and H.~Wen, ``Vidloc: A deep
  spatio-temporal model for 6-dof video-clip relocalization,'' in
  \emph{Proceedings of the IEEE Conference on Computer Vision and Pattern
  Recognition}, 2017, pp. 6856--6864.

\bibitem{melekhov2017image}
I.~Melekhov, J.~Ylioinas, J.~Kannala, and E.~Rahtu, ``Image-based localization
  using hourglass networks,'' in \emph{Proceedings of the IEEE international
  conference on computer vision workshops}, 2017, pp. 879--886.

\bibitem{walch2017image}
F.~Walch, C.~Hazirbas, L.~Leal-Taixe, T.~Sattler, S.~Hilsenbeck, and
  D.~Cremers, ``Image-based localization using lstms for structured feature
  correlation,'' in \emph{Proceedings of the IEEE International Conference on
  Computer Vision}, 2017, pp. 627--637.

\bibitem{sattler2016efficient}
T.~Sattler, B.~Leibe, and L.~Kobbelt, ``Efficient \& effective prioritized
  matching for large-scale image-based localization,'' \emph{IEEE transactions
  on pattern analysis and machine intelligence}, vol.~39, no.~9, pp.
  1744--1756, 2016.

\bibitem{schonberger2016structure}
J.~L. Schonberger and J.-M. Frahm, ``Structure-from-motion revisited,'' in
  \emph{Proceedings of the IEEE conference on computer vision and pattern
  recognition}, 2016, pp. 4104--4113.

\bibitem{sarlin2019coarse}
P.-E. Sarlin, C.~Cadena, R.~Siegwart, and M.~Dymczyk, ``From coarse to fine:
  Robust hierarchical localization at large scale,'' in \emph{Proceedings of
  the IEEE/CVF Conference on Computer Vision and Pattern Recognition}, 2019,
  pp. 12\,716--12\,725.

\bibitem{forstner1987fast}
W.~F{\"o}rstner and E.~G{\"u}lch, ``A fast operator for detection and precise
  location of distinct points, corners and centres of circular features,'' in
  \emph{Proc. ISPRS intercommission conference on fast processing of
  photogrammetric data}, vol.~6.\hskip 1em plus 0.5em minus 0.4em\relax
  Interlaken, 1987, pp. 281--305.

\bibitem{harris1988combined}
C.~Harris, M.~Stephens \emph{et~al.}, ``A combined corner and edge detector,''
  in \emph{Alvey vision conference}, vol.~15, no.~50.\hskip 1em plus 0.5em
  minus 0.4em\relax Citeseer, 1988, pp. 10--5244.

\bibitem{ng2003sift}
P.~C. Ng and S.~Henikoff, ``Sift: Predicting amino acid changes that affect
  protein function,'' \emph{Nucleic acids research}, vol.~31, no.~13, pp.
  3812--3814, 2003.

\bibitem{rublee2011orb}
E.~Rublee, V.~Rabaud, K.~Konolige, and G.~Bradski, ``Orb: An efficient
  alternative to sift or surf,'' in \emph{2011 International conference on
  computer vision}.\hskip 1em plus 0.5em minus 0.4em\relax Ieee, 2011, pp.
  2564--2571.

\bibitem{juan2009comparison}
L.~Juan and O.~Gwun, ``A comparison of sift, pca-sift and surf,''
  \emph{International Journal of Image Processing (IJIP)}, vol.~3, no.~4, pp.
  143--152, 2009.

\bibitem{dong2015domain}
J.~Dong and S.~Soatto, ``Domain-size pooling in local descriptors: Dsp-sift,''
  in \emph{Proceedings of the IEEE conference on computer vision and pattern
  recognition}, 2015, pp. 5097--5106.

\bibitem{bay2006surf}
H.~Bay, T.~Tuytelaars, and L.~V. Gool, ``Surf: Speeded up robust features,'' in
  \emph{European conference on computer vision}.\hskip 1em plus 0.5em minus
  0.4em\relax Springer, 2006, pp. 404--417.

\bibitem{tola2009daisy}
E.~Tola, V.~Lepetit, and P.~Fua, ``Daisy: An efficient dense descriptor applied
  to wide-baseline stereo,'' \emph{IEEE transactions on pattern analysis and
  machine intelligence}, vol.~32, no.~5, pp. 815--830, 2009.

\bibitem{rosten2006machine}
E.~Rosten and T.~Drummond, ``Machine learning for high-speed corner
  detection,'' in \emph{European conference on computer vision}.\hskip 1em plus
  0.5em minus 0.4em\relax Springer, 2006, pp. 430--443.

\bibitem{detone2018superpoint}
D.~DeTone, T.~Malisiewicz, and A.~Rabinovich, ``Superpoint: Self-supervised
  interest point detection and description,'' in \emph{Proceedings of the IEEE
  conference on computer vision and pattern recognition workshops}, 2018, pp.
  224--236.

\bibitem{yi2016lift}
K.~M. Yi, E.~Trulls, V.~Lepetit, and P.~Fua, ``Lift: Learned invariant feature
  transform,'' in \emph{European conference on computer vision}.\hskip 1em plus
  0.5em minus 0.4em\relax Springer, 2016, pp. 467--483.

\bibitem{savinov2017quad}
N.~Savinov, A.~Seki, L.~Ladicky, T.~Sattler, and M.~Pollefeys, ``Quad-networks:
  unsupervised learning to rank for interest point detection,'' in
  \emph{Proceedings of the IEEE conference on computer vision and pattern
  recognition}, 2017, pp. 1822--1830.

\bibitem{hartmann2014predicting}
W.~Hartmann, M.~Havlena, and K.~Schindler, ``Predicting matchability,'' in
  \emph{Proceedings of the IEEE conference on computer vision and pattern
  recognition}, 2014, pp. 9--16.

\bibitem{van2018efficient}
D.~Van~Opdenbosch, T.~Aykut, N.~Alt, and E.~Steinbach, ``Efficient map
  compression for collaborative visual slam,'' in \emph{2018 IEEE winter
  conference on applications of computer vision (WACV)}.\hskip 1em plus 0.5em
  minus 0.4em\relax IEEE, 2018, pp. 992--1000.

\bibitem{mera2020efficient}
M.~Mera-Trujillo, B.~Smith, and V.~Fragoso, ``Efficient scene compression for
  visual-based localization,'' in \emph{2020 International Conference on 3D
  Vision (3DV)}.\hskip 1em plus 0.5em minus 0.4em\relax IEEE, 2020, pp. 1--10.

\bibitem{camposeco2019hybrid}
F.~Camposeco, A.~Cohen, M.~Pollefeys, and T.~Sattler, ``Hybrid scene
  compression for visual localization,'' in \emph{Proceedings of the IEEE/CVF
  Conference on Computer Vision and Pattern Recognition}, 2019, pp. 7653--7662.

\bibitem{brachmann2018learning}
E.~Brachmann and C.~Rother, ``Learning less is more-6d camera localization via
  3d surface regression,'' in \emph{Proceedings of the IEEE Conference on
  Computer Vision and Pattern Recognition}, 2018, pp. 4654--4662.

\bibitem{balntas2017hpatches}
V.~Balntas, K.~Lenc, A.~Vedaldi, and K.~Mikolajczyk, ``Hpatches: A benchmark
  and evaluation of handcrafted and learned local descriptors,'' in
  \emph{Proceedings of the IEEE conference on computer vision and pattern
  recognition}, 2017, pp. 5173--5182.

\bibitem{revaud2019r2d2}
J.~Revaud, P.~Weinzaepfel, C.~De~Souza, N.~Pion, G.~Csurka, Y.~Cabon, and
  M.~Humenberger, ``R2d2: repeatable and reliable detector and descriptor,''
  \emph{arXiv preprint arXiv:1906.06195}, 2019.

\bibitem{dusmanu2019d2}
M.~Dusmanu, I.~Rocco, T.~Pajdla, M.~Pollefeys, J.~Sivic, A.~Torii, and
  T.~Sattler, ``D2-net: A trainable cnn for joint detection and description of
  local features,'' \emph{arXiv preprint arXiv:1905.03561}, 2019.

\bibitem{ono2018lf}
Y.~Ono, E.~Trulls, P.~Fua, and K.~M. Yi, ``Lf-net: Learning local features from
  images,'' \emph{Advances in neural information processing systems}, vol.~31,
  2018.

\bibitem{mishchuk2017working}
A.~Mishchuk, D.~Mishkin, F.~Radenovic, and J.~Matas, ``Working hard to know
  your neighbor's margins: Local descriptor learning loss,'' \emph{Advances in
  neural information processing systems}, vol.~30, 2017.

\bibitem{Dusmanu2019CVPR}
M.~Dusmanu, I.~Rocco, T.~Pajdla, M.~Pollefeys, J.~Sivic, A.~Torii, and
  T.~Sattler, ``{D2-Net: A Trainable CNN for Joint Detection and Description of
  Local Features},'' in \emph{Proceedings of the 2019 IEEE/CVF Conference on
  Computer Vision and Pattern Recognition}, 2019.

\bibitem{shotton2013scene}
J.~Shotton, B.~Glocker, C.~Zach, S.~Izadi, A.~Criminisi, and A.~Fitzgibbon,
  ``Scene coordinate regression forests for camera relocalization in rgb-d
  images,'' in \emph{Proceedings of the IEEE Conference on Computer Vision and
  Pattern Recognition}, 2013, pp. 2930--2937.

\bibitem{arandjelovic2016netvlad}
R.~Arandjelovic, P.~Gronat, A.~Torii, T.~Pajdla, and J.~Sivic, ``Netvlad: Cnn
  architecture for weakly supervised place recognition,'' in \emph{Proceedings
  of the IEEE conference on computer vision and pattern recognition}, 2016, pp.
  5297--5307.

\end{thebibliography}

\end{document}